\documentclass[conference, letterpaper]{IEEEtran}
\IEEEoverridecommandlockouts
\usepackage[numbers, sort]{natbib}
\usepackage[utf8]{inputenc} 
\usepackage[T1]{fontenc}    
\usepackage{hyperref}       
\usepackage{url}            
\usepackage{booktabs}       
\usepackage{amsfonts}       
\usepackage{nicefrac}       
\usepackage{microtype}      
\usepackage{graphicx}
\usepackage{subcaption}
\usepackage{adjustbox}
\usepackage{amsmath,amssymb,amsfonts}
\usepackage[acronym]{glossaries}
\usepackage{comment}
\usepackage{xcolor}
\usepackage{enumerate}
\usepackage[capitalise]{cleveref}
\usepackage[labelsep=period, format=plain, justification=justified, singlelinecheck=false, font={small, stretch=0.80}]{caption}
\usepackage{tabularx}
\usepackage{svg}
\usepackage{listings}
\usepackage{siunitx}
\usepackage[section]{placeins}
\usepackage[english]{babel}
\usepackage{booktabs}
\usepackage{multirow}
\usepackage{cancel}
\usepackage[normalem]{ulem}
\def\BibTeX{{\rm B\kern-.05em{\sc i\kern-.025em b}\kern-.08em
    T\kern-.1667em\lower.7ex\hbox{E}\kern-.125emX}}
\newacronym{ros}{ROS}{Robot Operating System}
\newacronym{pvc}{PVC}{Polyvinyl Chloride}
\newacronym{mae}{MAE}{Mean Absolute Error}
\newacronym{lidar}{LiDAR}{Light Detection and Ranging}
\newacronym{radar}{RADAR}{Radio Detection and Ranging}
\newacronym{ml}{ML}{Machine Learning}
\newacronym{tof}{ToF}{Time of Flight}
\newacronym{nir}{NIR}{Near-Infrared Radiation}
\newacronym{ir}{IR}{Infrared Radiation}
\newacronym{fmcw}{FMCW}{Frequency Modulated Continuous Wave}
\newacronym{if}{IF}{Intermediate Frequency}
\newacronym{aoa}{AoA}{Angle of Arrival}
\newacronym{cfar}{CFAR}{Constant False Alarm Rate}
\newacronym{iot}{IoT}{Internet of Things}

\usepackage{eso-pic}

\newcommand\AtPageUpperMyleft[1]{\AtPageUpperLeft{%
\put(\LenToUnit{1cm},\LenToUnit{-2cm}){#1}%
}}%

\AddToShipoutPictureBG*{%
  \AtPageUpperMyleft{\parbox[b][2cm][c]{\paperwidth}{%
    \centering
    \fontsize{12}{14}\selectfont
    \color{gray!50}
    This paper has been accepted for publication at the\\
    IEEE 9th World Forum on Internet of Things, Aveiro 2023. \copyright{}IEEE
  }}%
}

\AddToShipoutPictureBG*{
  \AtPageLowerLeft{%
    \raisebox{25pt}{\makebox[\paperwidth]{\begin{minipage}{21cm}\centering
    \fontsize{10}{12}\selectfont
      \textcolor{gray!50}{ \copyright 2023 IEEE.  Personal use of this material is permitted.  Permission from IEEE must be obtained for all other uses, in any current or future media, including reprinting/republishing this material for advertising or promotional purposes, creating new collective works, for resale or redistribution to servers or lists, or reuse of any copyrighted component of this work in other works.
      }
    \end{minipage}}}%
  }
}
    
\begin{document}

\title{Assessing the Robustness of LiDAR, Radar and Depth Cameras Against Ill-Reflecting Surfaces in Autonomous Vehicles: An Experimental Study\\
}

\author{\IEEEauthorblockN{Michael Lötscher, Nicolas Baumann, Edoardo Ghignone, Andrea Ronco, Michele Magno}
\IEEEauthorblockA{\textit{Department of Information Technology and Electrical Engineering, ETH Zurich} \\
Z\"urich, Switzerland \\
mloetscher@ethz.ch, \{nicolas.baumann, edoardo.ghignone, andrea.ronco, michele.magno\}@pbl.ee.ethz.ch} 
}

\maketitle

\begin{abstract}
Range-measuring sensors play a critical role in autonomous driving systems. While \gls{lidar} technology has been dominant, its vulnerability to adverse weather conditions is well-documented. This paper focuses on secondary adverse conditions – the implications of ill-reflective surfaces on range measurement sensors. We assess the influence of this condition on the three primary ranging modalities used in autonomous mobile robotics: \gls{lidar}, \gls{radar}, and Depth-Camera. Based on accurate experimental evaluation the paper's findings reveal that under ill-reflectivity, \gls{lidar} ranging performance drops significantly to 33\% of its nominal operating conditions, whereas \gls{radar} and Depth-Cameras maintain up to 100\% of their nominal distance ranging capabilities. Additionally, we demonstrate on a 1:10 scaled autonomous racecar how ill-reflectivity adversely impacts downstream robotics tasks, highlighting the necessity for robust range sensing in autonomous driving.
\end{abstract}


\section{Introduction}

In the context of today's \gls{iot}-driven world, where autonomous agents are becoming increasingly integrated into various aspects of daily life, from self-driving cars \cite{AD_smartcities} to robotic platforms for retirement communities \cite{elderly_kart}, and collaborative drones in smart cities \cite{drone_smartcities}, evaluating the robustness of these autonomous robot systems is paramount. As such, precise and robust range measurements play a critical role in ensuring the correct operation of the majority of autonomous mobile robotic systems, serving as a fundamental element in many mapping and localization algorithms \cite{ cartographer, MIT_PF}.

This principle is particularly significant in the realm of autonomous driving, where mapping and localization function as vital elements in the core robotics tasks of planning \cite{frazzoliRRT, linigerMPCC} and subsequent control strategies \cite{linigerMPCC, linigerVIA}, directly influencing the safety of roads. As highlighted for the case of \gls{radar} sensors in \cite{radar_int}, ensuring the accuracy and reliability of these measurements is critical for the effective functioning of autonomous vehicles, contributing to enhanced road safety and overall public well-being. Consequently, the importance of range measurements cannot be overstated, as they constitute one of the primary exteroceptive modalities in the considered systems \cite{cartographer, FengHaase2020deep}.  

\begin{figure}[]
    \centering
    \includegraphics[width=\columnwidth]{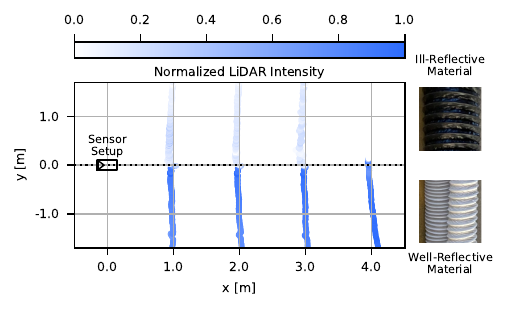}
     \caption{A visualization of how ill-reflective surfaces perturb the \gls{lidar} range measurements at different distances. The plot has been rescaled in order to align the sensor setup (on the left) with the axes. On the top half the range measurements are taken on ill-reflective material, and on the bottom one on well-reflective material. A sample of the corresponding surface is also shown on the right.}
    \label{fig:bbtvswhite}
\end{figure}


Among the variety of sensors utilized in autonomous driving, \gls{lidar} is a dominant player due to its high range and accuracy in distance measurements \cite{FengHaase2020deep}. 
This is reflected by \gls{lidar}'s prevalent use in many autonomous driving \gls{ml} datasets: when considering the survey by Feng et al. \cite{FengHaase2020deep}, 86\% of the datasets (18 out of 21) contain \gls{lidar} data.

However, despite its wide adoption, recent studies reveal that \gls{lidar} sensors, although superior under nominal conditions, are susceptible to adverse conditions \cite{aw_survey, hrfuser, aw_betz_radarcamera, aw_radarcamera_survey}. This susceptibility implies that \gls{lidar} sensor readings are not entirely reliable under adverse conditions, which can, in turn, affect downstream robotic tasks in autonomous driving due to inaccurate range measurements. Therefore, it is crucial to road safety and the wider acceptance of autonomous driving, to ensure robust and reliable range sensor readings \cite{cartographer}.

In contrast to previous robustness evaluations of autonomous driving sensor modalities, that primarily concentrate on adverse weather conditions \cite{aw_survey} or explicitly adversarial attackers \cite{ad_attacks}, this work focuses on a second fundamental and pivotal concern: the challenge of ill-reflecting surfaces. As discussed later in \cref{sec:background}, the principle of active exteroceptive \gls{tof} sensors such as \emph{\gls{lidar}} and \emph{\gls{radar}} is fundamentally linked to the reflectivity properties of the objects being measured. For instance, diffuse and dark-colored objects reflect significantly less signal, thus proving more challenging for range measurement sensors to detect, as demonstrated in \cref{fig:bbtvswhite}, where it can be seen that the diffuse and black-colored tubes are factually invisible to the \gls{lidar} starting already at a close range. Therefore, ill-reflecting objects, such as a dark-painted car with a matt paint finish, pose a common and great challenge to autonomous driving.

Hence, this study investigates the robustness with respect to ill-reflecting surfaces of the three most widely used range measurement sensors in mobile robotics: \emph{\gls{lidar}}, \emph{\gls{radar}} and \emph{Depth-Camera}. 
Furthermore, we explore their impact on downstream robotics tasks, particularly the localization of an autonomous racing car. Our study utilizes a 1:10 scaled autonomous racing platform known as F1TENTH \cite{f110, okelly2020f1tenth}. This platform facilitates safe, cost-effective, and rapid prototyping and has been instrumental in illustrating the principles and fundamentals of autonomous driving in various robotic tasks such as perception \cite{MIT_PF, iwasi}, planning \cite{formulazero} and control \cite{MAP, evansTCDriverClone}.

\section{Background} \label{sec:background}
In the realm of autonomous navigation, range sensors are instrumental in providing an understanding of the environment by creating a point cloud representation. This offers the necessary data required for tasks such as object detection, mapping, and localization. The following section presents the operating principles of the employed sensors in this study.

\subsection{LiDAR}
\gls{lidar} is a \gls{tof} sensor utilizing laser light in the \gls{nir} regime, hence a range estimate is produced based on the time needed for the light to reach the measurement target and return. It is comprised of a transmitter, which directs a collimated beam onto an object. This beam’s trajectory is usually manipulated by a mechanical mechanism, utilizing a rotating mirror to sweep the beam across a specific scene for comprehensive coverage. A receiver then detects the portion of the reflected light, that aligns closely with the path of emitted beam \cite{Siegwart2011}.

This study examined the \emph{Hokuyo UST-10LX}, which scans the environment in a 2D plane with a sampling rate of 40~Hz and a range of 10~m \cite{datasheet:Hokuyo}, thus making it ideal for applications in scaled mobile robotics and autonomous driving.

\subsection{\gls{fmcw} \gls{radar}}
\gls{radar}s sense the surroundings by illuminating the target with an electromagnetic signal and measuring the properties of the reflected (echo) signal. Different \gls{radar} categories are defined based on the illumination signal's properties.

\gls{fmcw} \gls{radar}s utilize frequency modulation to resolve the position and the relative velocity of the targets simultaneously, which makes them a favorite choice in robotics and automotive applications.
They operate on the baseband signal, also called \textit{Beat} or \gls{if} signal, which is obtained by mixing the transmitted signal with the received echo.
Important characteristics of \gls{fmcw} radar systems are the carrier frequency $f_c$, the chirp bandwidth $B$, and the sampling rate of the \gls{if} signal $f_s$.

The distance of the target can be resolved in the frequency domain, as it is directly proportional to the frequency of the beat signal, with a distance resolution of $\frac{c}{2B}$ where $c$ is the speed of light.
Once the object is resolved in distance, its velocity can be estimated by measuring the phase difference of two consecutive chirps at the corresponding range. The speed resolution is $\frac{\lambda}{4T_s}$, where $\lambda$ is the wavelength of the carrier frequency, and $T_s$ is the time resolution on the \gls{if} signal. In practice, more than two chirps are necessary to separate different targets at the same range properly.
When multiple antennas are available, it is also possible to estimate the \gls{aoa} of the signal through the process of digital beamforming. This step, combined with a \gls{cfar} filter, is essential for generating \gls{radar} point clouds. 

In this paper we evaluated the \emph{Texas Instruments IWR1443}, which operates in the frequency range \SI{76}{\giga\hertz}-\SI{81}{\giga\hertz} and offers a ranging rate of \SI{30}{\hertz}, with a maximum TX power of \SI{12}{\deci Bm}.

\subsection{Depth-Camera}
In the context of this paper, a Depth-Camera, specifically an \emph{Intel Realsense D455}, serves as an effective tool for the 3D reconstruction of the scene by creating a point cloud from the calculated distances. The camera achieves range measurements up to a distance of 6~m with a framerate of up to 90~Hz, at a resolution of $1280\times720$ pixel with a global shutter sensor \cite{datasheet:Realsense}.

The operational mechanics of this Depth-Camera incorporate an \gls{ir} stereo module. Its camera projects infrared light onto the scene, illuminating it and capturing the reflected portion. Two cameras, situated at a known distance apart from each other, a measure, referred to as the baseline $b$, capture the scene from slightly different perspectives, similar to how human vision functions. 

In this stereo vision setup, the disparity $d$ plays a vital role. Disparity refers to the difference in the apparent position of an object seen from the two cameras' perspectives. It is directly proportional to the distance between the object and the camera setup. Thus, by calculating the disparity, the depth $z$ of each point in the scene can be derived, as shown in \cref{eq:dc_dist}.

\begin{equation}
    \text{z} = \frac{bf}{d}
    \label{eq:dc_dist}
\end{equation}

The focal length $f$, is an intrinsic camera parameter and can be obtained by employing camera calibration techniques \cite{kalibr}, which assess the capability of the camera lens to focus light.

\section{Experimental Setup} \label{sec:exp_setup}

This section provides comprehensive information on the experimental setup designed to assess the robustness of different sensor modalities' range measurements. The evaluation is organized into three primary components: reflectivity characteristics measurements of the boundary materials, static range measurements to determine the data quality under varying reflectivity conditions, and dynamic range measurements to analyze the impact of range measurement quality on subsequent robotic tasks.

\subsection{Material Evaluation}
The materials selected for this analysis were exhaust tubes made from \gls{pvc}, chosen for their unique color, surface texture, and reflectivity attributes. The materials include a well-reflecting, white-colored, matt material serving as a baseline, and a black-colored, diffuse material with low reflectivity to examine sensor performance under challenging conditions. The material in question can be seen in \cref{fig:bbtvswhite}.

\begin{figure}[htbp]
    \centering
    \includegraphics[width=\columnwidth]{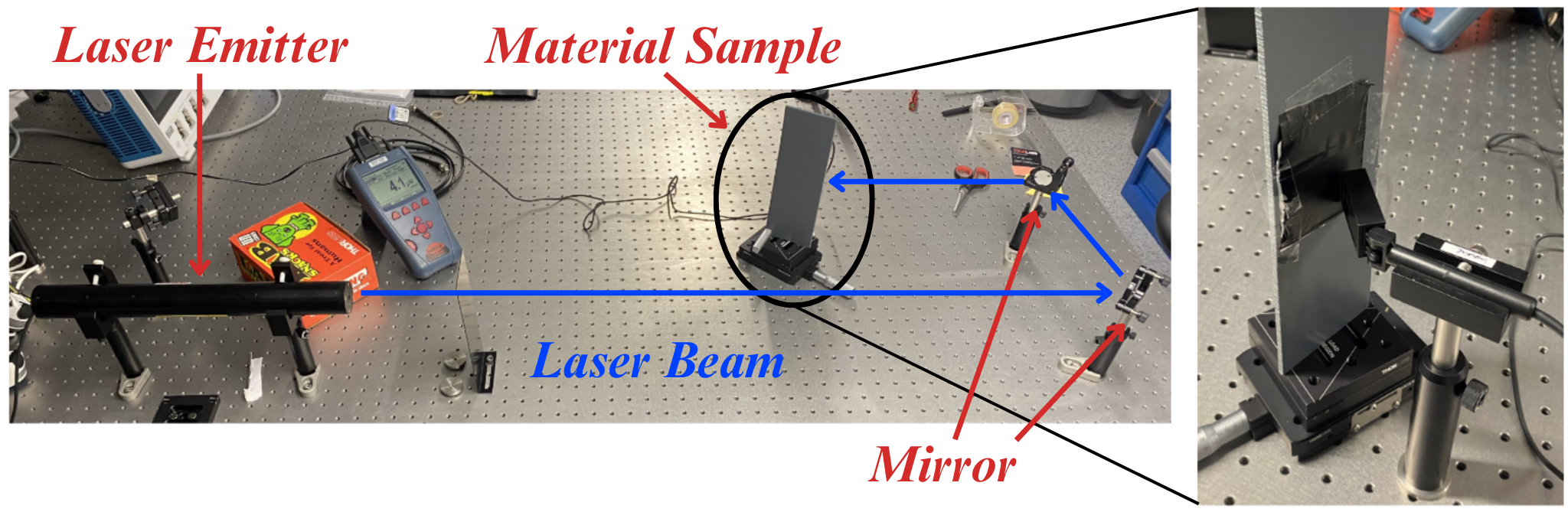}
    \caption{Setup to determine the power reflectivity $R$ by measuring the reflected power of the target material. The emitted laser beam is directed onto the sample via two mirrors and reflected subsequently at the target sample, whereupon the reflected component is measured by the detector.}
    \label{fig:refl_setup}
\end{figure}

To characterize the power reflectivity $R$ of these materials, an experiment was set up, as in \cref{fig:refl_setup}, using a helium-neon laser, which emits continuous waves with a power of 5~mW at a wavelength of 632.8~nm. An \emph{OPHIR PD300 UV} wide spectral range photodiode detector was used to measure the intensity of the reflected beams. The power reflectivity of the materials was recorded at twelve different locations on the sample, for both s- and p-polarization states.

\subsection{Static Experiment}
\begin{figure}[htbp]
    \centering
    \includegraphics[width=0.8\columnwidth]{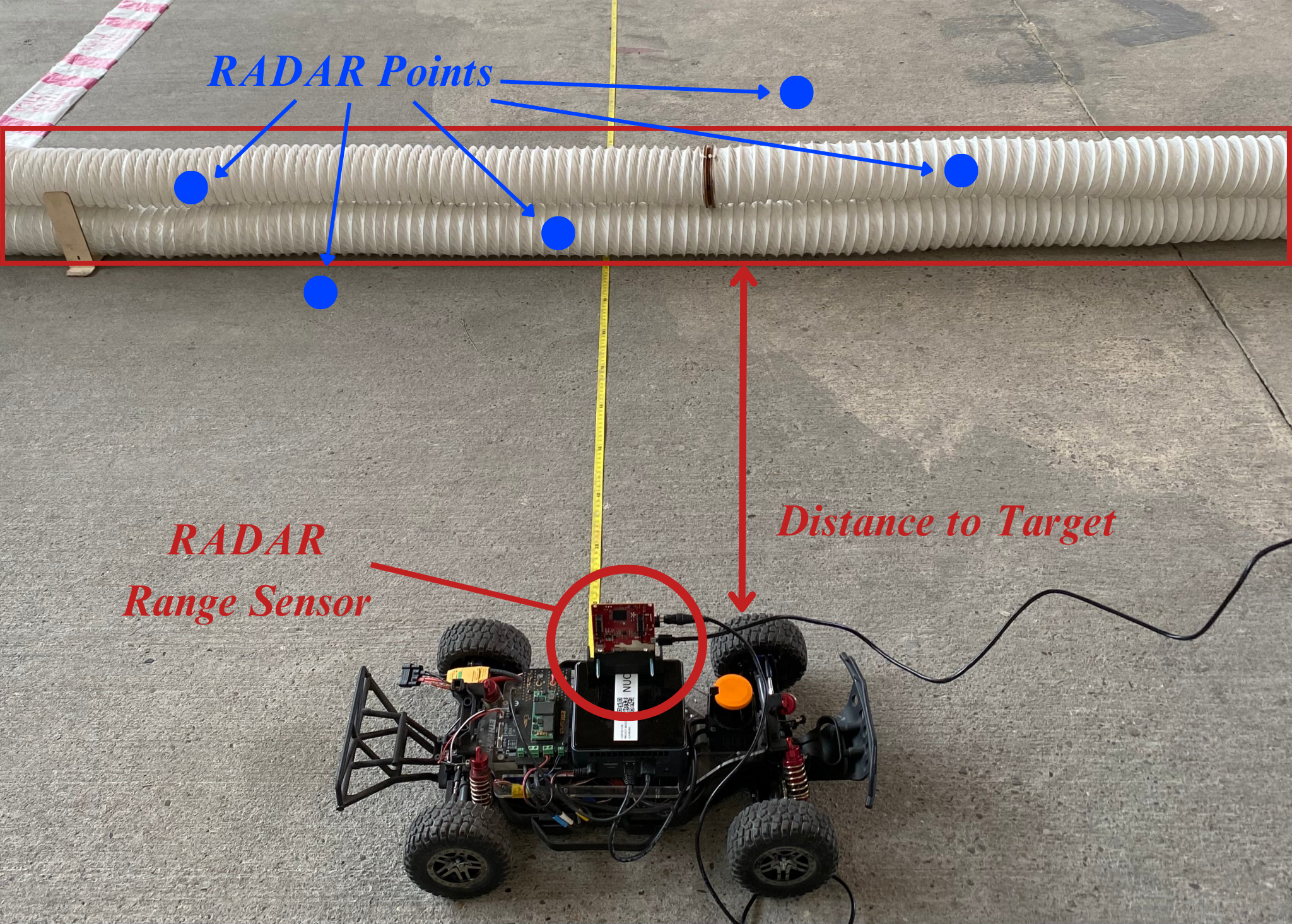}
    \caption{Experimental setup for the static performance assessment of the sensors, with the \gls{radar} sensor as an example. Blue \gls{radar} points are illustrated as a qualitative visualization of range measurements.}
    \label{fig:static_setup}
\end{figure}

For the static range measurements, a setup, similar to the one described by Cooper et al. \cite{Cooper2018}, was used. The sensor was positioned at a fixed distance from the tube, with data collection taking place for a consistent duration of 10~s for each measurement. The gathered data points were then compared to the \emph{ground truth} determined through a \emph{Bosch PLR 40C} rangefinder and a rolling meter, to gauge the sensor's accuracy. An exemplary depiction of the static reflectivity setup can be seen in \cref{fig:static_setup}.

\subsection{Dynamic Experiment}
The dynamic range measurement experiment was conducted on an oval-shaped racetrack, designed to simulate a mix of ideal and adverse conditions, half of the track featured well-reflective boundaries while the other half had ill-reflective ones. Ten consecutive laps were driven to gather data for each sensor's output, which was then processed by the localization, planning, and control systems of the robotic stack. The average lap time was computed to evaluate the overall performance.

\begin{figure}[htbp]
    \centering
    \includegraphics[trim={0 60 0 0},clip,width=\columnwidth]{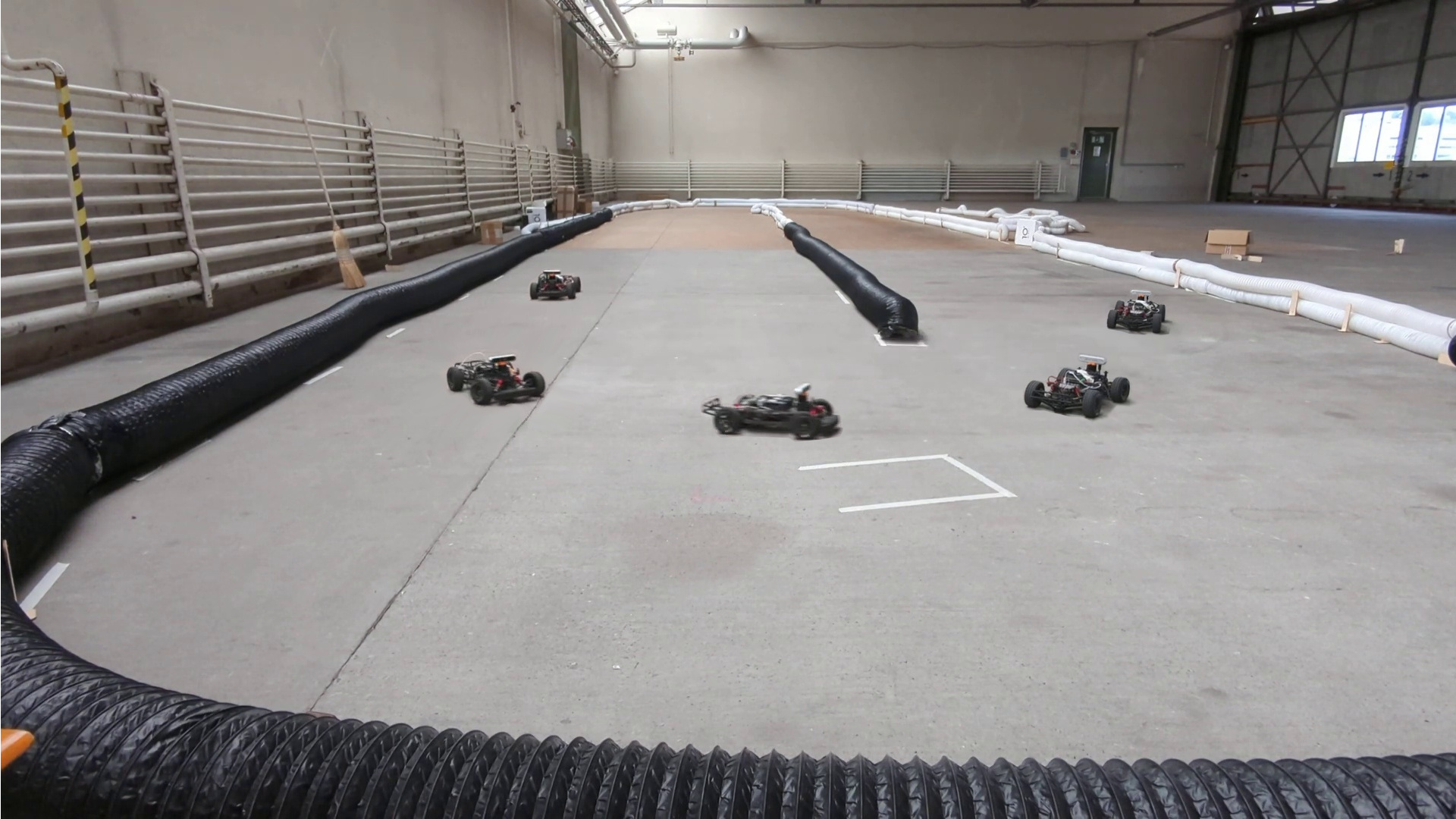}
    \caption{Oval-shaped racetrack setup designed for evaluating the lap times of various sensor configurations. The track is characterized by its distinct half-and-half boundary, composed of a black, ill-reflectivity sector and a white, well-reflecting one.}
    \label{fig:racetrack_laptime}
\end{figure}

A data-level sensor fusion approach was implemented to combine the data from the \gls{lidar}, \gls{radar}, and Depth-Camera sensors. The data from \gls{lidar} was given priority due to its superior performance in static experiments, while the Depth-Camera and \gls{radar} data were utilized when \gls{lidar} data was unavailable. The aim of this dynamic experiment was to understand how sensor configurations and range measurement quality can impact the racecar's behavior under varying environmental conditions.

\section{Performance Evaluation Parameters}
The \gls{mae} and standard deviation performance parameters are employed in this paper. They quantify the accuracy and variability of the sensor measurements. The \gls{mae} provides a measure of the average magnitude of error in a set of predictions, without considering their direction. It is calculated as the average over the verification sample of the absolute values of the differences between the forecast and the corresponding observation, as shown in \cref{eq:mae}.

\begin{equation}
    \text{\gls{mae}} = \frac{1}{N} \sum_{i=1}^{N} \left| y_i - \hat{y}_i \right|
    \label{eq:mae}
\end{equation}

In the equation shown, $y_i$ denotes the ground truth range, while $\hat{y}_i$ denotes the range measured by the sensors. $N$ is the total number of samples. The standard deviation, on the other hand, measures the dispersion of the measured data. It is used to quantify the amount of variation in a set of values. A low standard deviation indicates that the values tend to be close to the mean of the data, while a high standard deviation indicates that the values are spread out over a wider range. The formula of the standard deviation is described in \cref{eq:sd}.

\begin{equation}
    \sigma = \sqrt{\frac{1}{N} \sum_{i=1}^{N} \left( y_i - \bar{y} \right)^2}
    \label{eq:sd}
\end{equation}

The mean of all sensor readings is denoted by $\bar{y}$, while each individual sensor reading is indicated by $y_i$. $N$ stands for the total number of samples. A lower value for the standard deviation $\sigma$ is more desirable, as it suggests a lower variation across the measurements.

\section{Results}
The results of the experiments described in \cref{sec:exp_setup} are presented and discussed in this section.

\subsection{Boundary Reflectivity Measurements}
The experimental data detailed in \cref{table:refl_data} highlights the contrasting reflectivity characteristics of the black and white colored materials under varying polarization conditions and their implications on range sensor performance. The power reflectivity $R$ is the divisor of the incident and reflected power, as seen in \cref{eq:R}. The incident beam power $P$$_{i}$ was measured as 4.63~mW.

\begin{equation}
    R = \frac{P_r}{P_i}
    \label{eq:R}
\end{equation}

The standard deviation of $R$ is calculated using Gaussian error propagation under the assumption, that the standard deviation of the incident beam power $\sigma_{P_i}$ can be neglected and therefore simplifies to \cref{eq:sigma_R}.

\begin{equation}
    \sigma_R = \frac{\sigma_{P_r}}{P_i}
    \label{eq:sigma_R}
\end{equation}

The reflected power $P_{r}$ and the power reflectivity $R$ of the white-colored material surpass those of the black-colored material for both p-polarized and s-polarized incident light. For the white-colored material, the maximum observed power reflectivity reached  9.5\%, in stark contrast with the much lower 4.7\% and 0.3\% recorded for the black-colored material under s- and p-polarized light respectively. Although higher than the power reflectivities seen for the black-colored material, the white material is still within a relatively low reflectivity range. Interestingly, the power reflectivity of the black-colored material is 13 times lower when exposed to a p-polarized incident laser compared to an s-polarized laser, marking a significant difference. 

The distinct reflectivity characteristics of the white-colored and black-colored materials express a clear comparative advantage for the white material. These properties serve as an optimal test bed for assessing sensor performance across a range of conditions. The relatively higher reflectivity of the white-colored material provides a baseline for standard environments, while the ill-reflectivity of the black-colored material allows for the examination of sensor performance under more challenging, adverse conditions. This range of reflectivities facilitates investigations into optimizing sensor configurations to tune effectively across diverse settings. \\

\begin{table}[!htbp]
    \caption{Reflectivity properties of ill-reflective black-colored boundaries and well-reflective white-colored boundaries. Mean reflected power $P_{r}$ in $\mu W$; Standard deviation $\sigma_{Pr}$ in $\mu W$; Power reflectivity $R$ (dimensionless); Standard deviation $\sigma_{R}$ (dimensionless).}
    \centering
    \renewcommand{\arraystretch}{1.35}
    \begin{tabular}{@{}l*{4}{c}c@{}}
    \toprule
    \multirow{2}{*}{} & \multicolumn{2}{c}{Black mat.} & \multicolumn{2}{c}{White mat.}  \\
    & p-pol. & s-pol. & s-pol. & p-pol.  \\
    \midrule
    $P_{r}$ [$\mu$W] & 14.08 & 219.63 & 367.42 & \textbf{440.75} \\
    $\sigma_{Pr}$ [$\mu$W] & \textbf{1.70} & 25.54 & 54.79 & 76.70 \\ 
    $R$ [$\cdot10^{-3}$]& 3.041 & 47.436 & 79.356 & \textbf{95.194} \\
    $\sigma_{R}$ [$\cdot10^{-3}$] & \textbf{0.367} & 5.516 & 11.834 & 16.566 \\
    \bottomrule
    \end{tabular}
    \label{table:refl_data}
\end{table}

\subsection{Static Range Measurements}
\begin{figure}[htbp]
    \centering
    \includegraphics[width=1\linewidth,trim={0 0.2cm 0 0.2cm},clip]{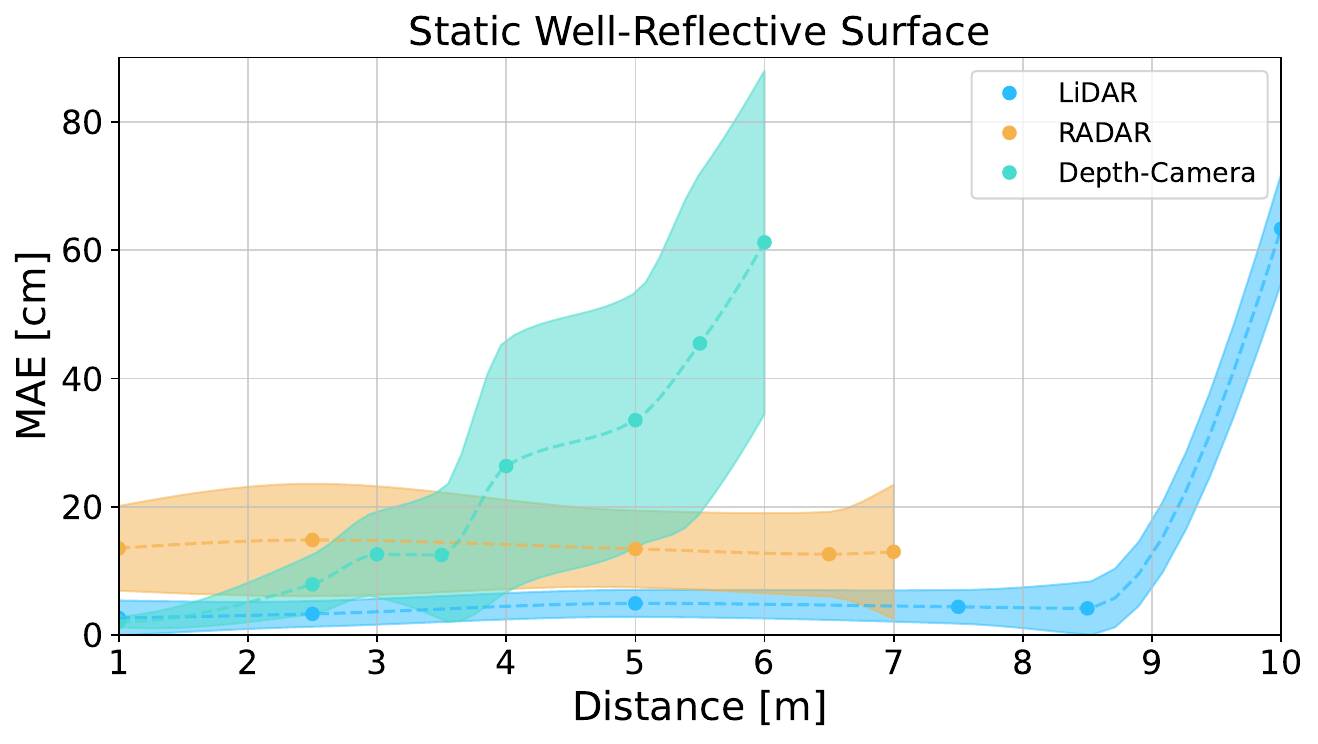}
    \caption{Comparsion of \gls{lidar}, \gls{radar} and Depth-Camera sensor in terms of range and \gls{mae}, when sensing \emph{well-reflective} boundaries. The standard deviation of the measurements is represented by the shaded regions. Lower is better.}
    \label{fig:static_wt}
\end{figure}

The outcomes of the static experiment conducted on the well-reflective baseline material, are illustrated  in \cref{fig:static_wt}. The \gls{lidar} sensor exhibits the most extensive range for this material, reaching its maximum range of 10~m, achieving the manufacturer's specified maximum range \cite{datasheet:Hokuyo}. This surpasses the \gls{radar} sensor's maximum range of 7~m and the Depth-Camera's limit of 6~m. The \gls{lidar}'s range measurements maintain a \gls{mae} that is bound below 5~cm up to a distance of 8.5~m, however growing drastically beyond this mark. 

The \gls{radar} sensor's performance is comparable to the \gls{lidar}'s performance. However, it achieves only a maximum range of 7~m and shows a mostly constant but larger \gls{mae} between 10 to 15~cm.

Contrary to the consistent accuracy of the \gls{lidar} and \gls{radar} sensor, the accuracy of the Depth-Camera's measurements is monotonically decreasing as the distance from the target increases. This trend is witnessed in both the \gls{mae} as well as the standard deviation. Consequently, the \gls{mae} is resulting in 61~cm at a distance of 6~m to the Depth-Camera. \\
    
\begin{figure}[htbp]
    \centering
    \includegraphics[width=\linewidth,trim={0 0.2cm 0 0.2cm},clip]{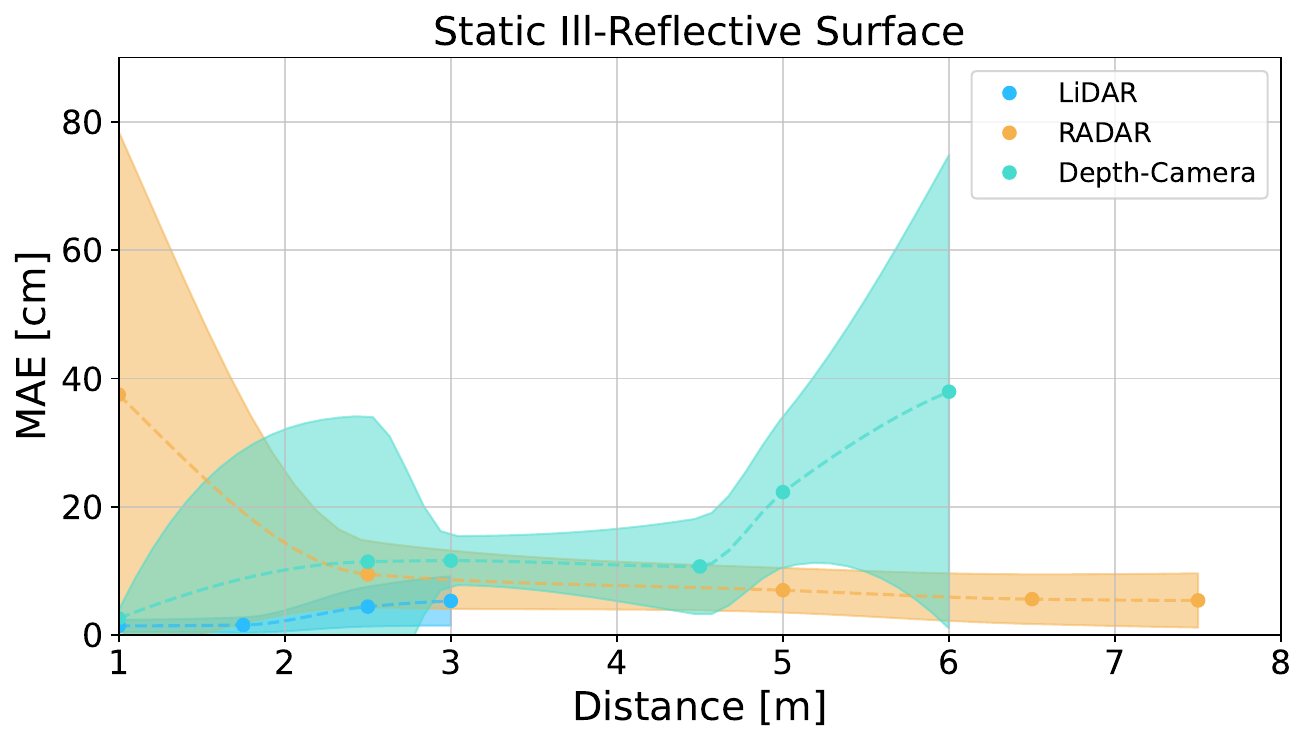}
    \caption{Comparsion of \gls{lidar}, \gls{radar} and Depth-Camera sensor in terms of range and \gls{mae}, when sensing an \emph{ill-reflecting} surface. The standard deviation of the measurements is represented by the shaded regions. Lower is better.}
    \label{fig:static_bt}
\end{figure}

In \cref{fig:static_bt}, the static experiment results for the ill-reflecting surface are shown. The \gls{lidar} sensor's behavior is similar as for the other material shown in \cref{fig:static_wt}, maintaining a low and mostly consistent \gls{mae} and standard deviation throughout. However, its maximum range is significantly lower, achieving at most a range of 3~m. Therefore, we see that now the maximum ranging distance has been reduced to 33\% of the baseline sensing capabilities. 

The Depth-Camera sensor replicates its performance pattern observed with the baseline material, exhibiting the same maximum range and upward \gls{mae} trend. However, the actual error of the \gls{mae} is significantly lower at distances from the target to the sensor higher than 3~m.

An almost constant \gls{mae}, as well as standard deviation, is achieved by the \gls{radar} when measuring from a distance of 2.5~m onwards. The sensor manages to keep the \gls{mae} remaining below an error of 10~cm for this range. Notably, the maximum range of the \gls{radar} is 7.5~m, the highest of all sensors for the ill-reflective surface. Despite these strengths, an observable anomaly occurs at shorter distances, where false positives appear more frequently, causing both \gls{mae} and standard deviation to surge compared to their values at larger distances. Overall, the \gls{radar} outperforms the \gls{lidar} and Depth-Camera counterparts in range and accuracy for distances larger than 2.5~m for the ill-reflective material. \\

\subsection{Dynamic Range Measurements}

\begin{figure}[htbp]
    \centering
    \includegraphics[width=0.95\columnwidth,trim={0 0.2 0 1.2cm},clip]{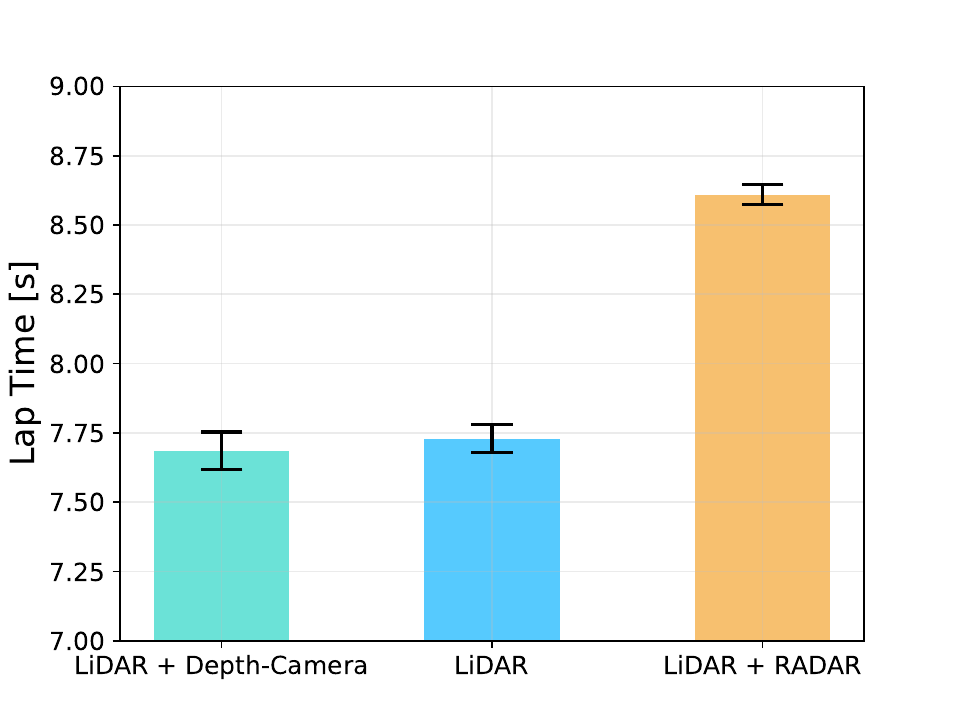}
    \caption{Comparsion of the mean lap times achieved by the \gls{lidar}-only with 7.729~s, \gls{lidar} + Depth-Camera Fusion with 7.686~s and \gls{lidar} + \gls{radar} with 8.011~s, sensor fusion configurations. Lower is better.}
    \label{fig:dynamic}
\end{figure}

\cref{fig:dynamic} presents the mean lap times for each of the sensor combinations. The \gls{lidar} + Depth-Camera configuration achieves the fastest lap times with a mean lap time of 7.686~s. It outperforms both the \gls{lidar}-only configuration as well as the \gls{lidar} + \gls{radar} fusion. Specifically, the \gls{lidar}-only configuration achieves a laptime of 7.729~s and lags by 0.043~s, hence a 0.573\% deterioration, while the \gls{lidar} + \gls{radar} fusion's lap times are considerably slower with 8.011~s, hence 12.026\% relative to the \gls{lidar} + Depth-Camera setup. Notably, the lap times variance is well contained across all configurations, showing a maximum standard deviation of no more than 0.054~s for all three configurations. This deviation accounts for 0.068~s in the context of the \gls{lidar} + Depth-Camera fusion, attesting to the consistency of the performance across the trials.

\section{Conclusion}

This paper highlights the substantial influence of ill-reflective materials on the ranging capabilities of exteroceptive sensors, such as \gls{lidar}, \gls{radar}, and Depth-Cameras. By shedding light on the effects of ill-reflectivity on autonomous mobile robotic systems, particularly within the realm of autonomous driving, our study underscores the importance of robust and precise range sensing in enhancing road safety. As we embrace the advancements of \gls{iot} in shaping autonomous technologies, our research contributes to the broader goal of making roads safer and more reliable for all users. Hence, an assessment of static ranging abilities linked to these sensor modalities exposes that, although \gls{lidar} yields the highest and most consistent ranging capabilities under nominal conditions, these degrade to a mere 33\% under the adverse condition of ill-reflectivity. Conversely, other technologies such as \gls{radar} and Depth-Cameras significantly outperform \gls{lidar} in such scenarios, retaining full-ranging distance capabilities as under nominal conditions. Through an experimental evaluation on a 1:10 scale autonomous racing car, we demonstrate the consequential effect of the quality of ranging measurements on downstream robotics tasks. This underscores the importance of embracing multimodal ranging sensors and sensor fusion to ensure robust and reliable autonomous driving behavior.

\section*{Acknowledgments}
The authors wish to express their gratitude to the Ultrafast Dynamics Group of ETH Zurich, particularly Vladimir Ovuka, for granting access to their laboratories and for supporting the conduction of our reflectivity experiments.

\bstctlcite{IEEEexample:BSTcontrol} 
\bibliographystyle{IEEEtranDOI}
\bibliography{main}

\begin{thebibliography}{10}
\providecommand{\url}[1]{#1}
\csname url@samestyle\endcsname
\providecommand{\newblock}{\relax}
\providecommand{\bibinfo}[2]{#2}
\providecommand{\BIBentrySTDinterwordspacing}{\spaceskip=0pt\relax}
\providecommand{\BIBentryALTinterwordstretchfactor}{4}
\providecommand{\BIBentryALTinterwordspacing}{\spaceskip=\fontdimen2\font plus
\BIBentryALTinterwordstretchfactor\fontdimen3\font minus
  \fontdimen4\font\relax}
\providecommand{\BIBforeignlanguage}[2]{{%
\expandafter\ifx\csname l@#1\endcsname\relax
\typeout{** WARNING: IEEEtran.bst: No hyphenation pattern has been}%
\typeout{** loaded for the language `#1'. Using the pattern for}%
\typeout{** the default language instead.}%
\else
\language=\csname l@#1\endcsname
\fi
#2}}
\providecommand{\BIBdecl}{\relax}
\BIBdecl

\bibitem{AD_smartcities}
I.~Yaqoob, L.~U. Khan, S.~M.~A. Kazmi, M.~Imran, N.~Guizani, and C.~S. Hong,
  ``Autonomous driving cars in smart cities: Recent advances, requirements, and
  challenges,'' \emph{IEEE Network}, vol.~34, no.~1, pp. 174--181, 2020, doi:
  10.1109/MNET.2019.1900120.

\bibitem{elderly_kart}
S.~El-Tawab, N.~Sprague, and A.~Mufti, ``Autonomous vehicles: Building a
  test-bed prototype at a controlled environment,'' in \emph{2020 IEEE 6th
  World Forum on Internet of Things (WF-IoT)}, 2020, pp. 1--6, doi:
  10.1109/WF-IoT48130.2020.9221222.

\bibitem{drone_smartcities}
S.~H. Alsamhi, O.~Ma, M.~S. Ansari, and F.~A. Almalki, ``Survey on
  collaborative smart drones and internet of things for improving smartness of
  smart cities,'' \emph{IEEE Access}, vol.~7, pp. 128\,125--128\,152, 2019,
  doi: 10.1109/ACCESS.2019.2934998.

\bibitem{cartographer}
W.~Hess, D.~Kohler, H.~Rapp, and D.~Andor, ``Real-time loop closure in 2d lidar
  slam,'' in \emph{2016 IEEE International Conference on Robotics and
  Automation (ICRA)}, 2016, pp. 1271--1278.

\bibitem{MIT_PF}
C.~H. Walsh and S.~Karaman, ``Cddt: Fast approximate 2d ray casting for
  accelerated localization,'' in \emph{2018 IEEE International Conference on
  Robotics and Automation (ICRA)}, 2018, pp. 3677--3684, doi:
  10.1109/ICRA.2018.8460743.

\bibitem{frazzoliRRT}
S.~Karaman and E.~Frazzoli, ``Sampling-based algorithms for optimal motion
  planning,'' 2011.

\bibitem{linigerMPCC}
\BIBentryALTinterwordspacing
A.~Liniger, A.~Domahidi, and M.~Morari, ``Optimization-based autonomous racing
  of 1:43 scale {RC} cars,'' \emph{Optimal Control Applications and Methods},
  vol.~36, no.~5, pp. 628--647, jul 2014, doi: 10.1002/oca.2123.
\BIBentrySTDinterwordspacing

\bibitem{linigerVIA}
\BIBentryALTinterwordspacing
A.~Liniger and J.~Lygeros, ``Real-time control for autonomous racing based on
  viability theory,'' \emph{{IEEE} Transactions on Control Systems Technology},
  vol.~27, no.~2, pp. 464--478, mar 2019, doi: 10.1109/tcst.2017.2772903.
\BIBentrySTDinterwordspacing

\bibitem{radar_int}
Y.~Wang, Q.~Zhang, Z.~Wei, Y.~Lin, and Z.~Feng, ``Performance analysis of
  coordinated interference mitigation approach for automotive radar,''
  \emph{IEEE Internet of Things Journal}, vol.~10, no.~13, pp.
  11\,683--11\,695, 2023, doi: 10.1109/JIOT.2023.3244566.

\bibitem{FengHaase2020deep}
D.~Feng, C.~Haase-Sch{\"u}tz, L.~Rosenbaum, H.~Hertlein, C.~Glaeser, F.~Timm,
  W.~Wiesbeck, and K.~Dietmayer, ``Deep multi-modal object detection and
  semantic segmentation for autonomous driving: Datasets, methods, and
  challenges,'' \emph{IEEE Transactions on Intelligent Transportation Systems},
  2020.

\bibitem{aw_survey}
\BIBentryALTinterwordspacing
Y.~Zhang, A.~Carballo, H.~Yang, and K.~Takeda, ``Perception and sensing for
  autonomous vehicles under adverse weather conditions: A survey,''
  \emph{{ISPRS} Journal of Photogrammetry and Remote Sensing}, vol. 196, pp.
  146--177, feb 2023, doi: 10.1016/j.isprsjprs.2022.12.021.
\BIBentrySTDinterwordspacing

\bibitem{hrfuser}
T.~Broedermann, C.~Sakaridis, D.~Dai, and L.~V. Gool, ``Hrfuser: A
  multi-resolution sensor fusion architecture for 2d object detection,'' 2023.

\bibitem{aw_betz_radarcamera}
F.~Nobis, M.~Geisslinger, M.~Weber, J.~Betz, and M.~Lienkamp, ``A deep
  learning-based radar and camera sensor fusion architecture for object
  detection,'' 2020.

\bibitem{aw_radarcamera_survey}
S.~Yao, R.~Guan, X.~Huang, Z.~Li, X.~Sha, Y.~Yue, E.~G. Lim, H.~Seo, K.~L. Man,
  X.~Zhu, and Y.~Yue, ``Radar-camera fusion for object detection and semantic
  segmentation in autonomous driving: A comprehensive review,'' 2023.

\bibitem{ad_attacks}
M.~A. Hoque and R.~Hasan, ``Autonomous driving security: A comprehensive threat
  model of attacks and mitigation strategies,'' in \emph{2022 IEEE 8th World
  Forum on Internet of Things (WF-IoT)}, 2022, pp. 1--6, doi:
  10.1109/WF-IoT54382.2022.10152219.

\bibitem{f110}
M.~O’Kelly, H.~Zheng, A.~Jain, J.~Auckley, K.~Luong, and R.~Mangharam,
  ``Tunercar: A superoptimization toolchain for autonomous racing,'' in
  \emph{2020 IEEE International Conference on Robotics and Automation (ICRA)},
  2020, pp. 5356--5362, doi: 10.1109/ICRA40945.2020.9197080.

\bibitem{okelly2020f1tenth}
M.~O’Kelly, H.~Zheng, D.~Karthik, and R.~Mangharam, ``F1tenth: An open-source
  evaluation environment for continuous control and reinforcement learning,''
  in \emph{NeurIPS 2019 Competition and Demonstration Track}.\hskip 1em plus
  0.5em minus 0.4em\relax PMLR, 2020, pp. 77--89.

\bibitem{iwasi}
A.~Ronco, N.~Baumann, M.~Giordano, and M.~Magno, ``Towards robust velocity and
  position estimation of opponents for autonomous racing using low-power
  radar,'' in \emph{2023 9th International Workshop on Advances in Sensors and
  Interfaces (IWASI)}, 2023, pp. 21--26, doi: 10.1109/IWASI58316.2023.10164312.

\bibitem{formulazero}
A.~Sinha, M.~O'Kelly, H.~Zheng, R.~Mangharam, J.~Duchi, and R.~Tedrake,
  ``Formulazero: Distributionally robust online adaptation via offline
  population synthesis,'' in \emph{Proceedings of the 37th International
  Conference on Machine Learning}, ser. ICML'20, 2020.

\bibitem{MAP}
J.~Becker, N.~Imholz, L.~Schwarzenbach, E.~Ghignone, N.~Baumann, and M.~Magno,
  ``Model- and acceleration-based pursuit controller for high-performance
  autonomous racing,'' in \emph{2023 IEEE International Conference on Robotics
  and Automation (ICRA)}, 2023, pp. 5276--5283, doi:
  10.1109/ICRA48891.2023.10161472.

\bibitem{evansTCDriverClone}
B.~D. Evans, H.~A. Engelbrecht, and H.~W. Jordaan, ``High-speed autonomous
  racing using trajectory-aided deep reinforcement learning,'' \emph{IEEE
  Robotics and Automation Letters}, vol.~8, no.~9, pp. 5353--5359, 2023, doi:
  10.1109/LRA.2023.3295252.

\bibitem{Siegwart2011}
R.~Siegwart, I.~R. Nourbakhsh, and D.~Scaramuzza, \emph{Introduction to
  Autonomous Mobile Robots}.\hskip 1em plus 0.5em minus 0.4em\relax The MIT
  Press, 2011, ISBN 0262015358.

\bibitem{datasheet:Hokuyo}
\BIBentryALTinterwordspacing
Hokuyo, ``Lidar ust-10lx,'' [Online]. Available:
  \url{https://hokuyo-usa.com/products/lidar-obstacle-detection/ust-10lx}.
\BIBentrySTDinterwordspacing

\bibitem{datasheet:Realsense}
\BIBentryALTinterwordspacing
Intel, ``Realsense d455,'' [Online]. Available:
  \url{https://www.intelrealsense.com/depth-camera-d455/}.
\BIBentrySTDinterwordspacing

\bibitem{kalibr}
L.~Oth, P.~Furgale, L.~Kneip, and R.~Siegwart, ``Rolling shutter camera
  calibration,'' in \emph{Proceedings of the IEEE Conference on Computer Vision
  and Pattern Recognition (CVPR)}, June 2013.

\bibitem{Cooper2018}
\BIBentryALTinterwordspacing
M.~A. Cooper, J.~F. Raquet, and R.~Patton, ``Range information characterization
  of the hokuyo ust-20lx lidar sensor,'' \emph{Photonics}, vol.~5, no.~2, 2018.
\BIBentrySTDinterwordspacing

\end{thebibliography}
\end{document}